\documentclass[journal]{IEEEtran}
\usepackage{xcolor,soul,framed} %,caption
\usepackage{graphicx}
\usepackage{float}
\usepackage{subfigure}
\usepackage{graphicx}
\usepackage{adjustbox}
\usepackage{color}
\usepackage{algorithm}
\usepackage{algorithmic}
\usepackage{amsfonts}
\usepackage{mathtools}
\usepackage{hyperref}
\usepackage{doi}

\colorlet{shadecolor}{yellow}
\usepackage{array}
\usepackage{amsmath}
\DeclareMathOperator*{\argmax}{argmax} 
\DeclareMathOperator*{\argmin}{argmin} 
\usepackage{mdwmath}
\usepackage{mdwtab}
\usepackage{eqparbox}
\usepackage{url}

\begin{document}
% \bstctlcite{IEEEexample:BSTcontrol}
    \title{Adaptive Social Metaverse Streaming based on Federated Multi-Agent Deep Reinforcement Learning}
    \author{Zijian Long, Haopeng Wang, Haiwei Dong,~\IEEEmembership{Senior Member,~IEEE}, and Abdulmotaleb El Saddik,~\IEEEmembership{Fellow,~IEEE}}
% \author{Zijian Long, Haiwei Dong,~\IEEEmembership{Senior Member,~IEEE}, and Abdulmotaleb El Saddik,~\IEEEmembership{Fellow,~IEEE}
% \thanks{Manuscript received 2023.}
% \thanks{Zijian Long and Abdulmotaleb El Saddik are with the Multimedia Communications Research Laboratory (MCRLab), the School of Electrical Engineering and Computer Science, University of Ottawa, Ottawa, ON K1N 6N5, Canada (e-mail: \{zlong038, elsaddik\}@uOttawa.ca).}
% \thanks{Haiwei Dong is with Ottawa Research Center, Huawei Technologies Canada, Ottawa, ON K2K 3J1, Canada (e-mail: haiwei.dong@ieee.org).}}
\markboth{IEEE Transactions on Computational Social Systems}%
{Shell \MakeLowercase{\textit{et al.}}: A Sample Article Using IEEEtran.cls for IEEE Journals}

\IEEEpubid{0000--0000/00\$00.00~\copyright~2024 IEEE}


% ====================================================================
\maketitle

% === ABSTRACT ====================================================================
% =================================================================================
\begin{abstract}
%\boldmath

The social metaverse is a growing digital ecosystem that blends virtual and physical worlds. It allows users to interact socially, work, shop, and enjoy entertainment. However, privacy remains a major challenge, as immersive interactions require continuous collection of biometric and behavioral data. At the same time, ensuring high-quality, low-latency streaming is difficult due to the demands of real-time interaction, immersive rendering, and bandwidth optimization. To address these issues, we propose ASMS (Adaptive Social Metaverse Streaming), a novel streaming system based on Federated Multi-Agent Proximal Policy Optimization (F-MAPPO). ASMS leverages F-MAPPO, which integrates federated learning (FL) and deep reinforcement learning (DRL) to dynamically adjust streaming bit rates while preserving user privacy. Experimental results show that ASMS improves user experience by at least 14\% compared to existing streaming methods across various network conditions. Therefore, ASMS enhances the social metaverse experience by providing seamless and immersive streaming, even in dynamic and resource-constrained networks, while ensuring that sensitive user data remains on local devices.

\end{abstract}

% === KEYWORDS ====================================================================
% =================================================================================

% \fi
%
% For peer review papers, this IEEEtran command inserts a page break and
\IEEEpeerreviewmaketitle
% ====================================================================
% ====================================================================
% ====================================================================

% === I. INTRODUCTION =============================================================
% =================================================================================

\section{Introduction} \label{introduction}
The metaverse is seen as the next evolution of the Internet, offering a seamless digital space where users can meet, socialize, play games, and collaborate in immersive 3D environments \cite{wang2022survey}. As adoption grows, it has gained significant global attention. Gartner predicts that by 2026, 25\% of people will spend at least an hour per day in metaverse environments \cite{metaversemarket}.  This rapid development of the metaverse is driven by the integration of several advanced technologies: extended reality (XR) provides an immersive 3D experience by headsets; digital twins enable real-time virtual representations of physical entities \cite{yang2022optimizing}; mobile edge computing (MEC) brings powerful computing servers closer to users \cite{long2023human}; 6G provides ultra-reliable low-latency communication between users and servers in the network edge; artificial intelligence (AI) makes smart decisions in a variety of aspects such as resource allocation to improve user experience; blockchain plays an important role in protecting users' digital assets in the metaverse \cite{yao2022freedom}; 3D reconstruction allows for the creation of realistic virtual spaces. 

Undoubtedly, the metaverse will have a profound influence across all facets of existence, particularly within the social domain \cite{zhang2022towards}. The social metaverse deconstructs established social platforms and eliminates isolated social applications, facilitating extensive and dynamic social interactions among user-controlled avatars in the long-lasting social spaces \cite{wang2023social}. The inherent characteristics of openness and decentralization in the social metaverse are poised to significantly transform prevailing paradigms in social interactions, enriching the overall social experience and fostering innovation in social commerce \cite{hennig2023social}. However, to accurately interpret user interactions and enable responsive, personalized experiences in the social metaverse, it is essential to continuously monitor and collect extensive real-life data from users. This includes tracking critical biometric and behavioral data, such as eye movements, hand gestures, and voice commands. In many situations, users are unaware of the ongoing recording and analysis of their data, even including personal information, behavior, and communication data \cite{falchuk2018social}, thereby putting their privacy at risk in unforeseen ways.

Mobile XR devices, such as Augmented Reality (AR), Virtual Reality (VR), and Mixed Reality (MR) headsets, have become the most popular interfaces to the social metaverse today \cite{song2022distributed}. These standalone XR devices are usually equipped with independent processors by which users would experience immersive scenes while maintaining convenient mobility. However, the seamless rendering of high-resolution scenes of the social metaverse requires large computing resources and energy, which cannot be met by XR devices themselves. Thanks to the advent of MEC \cite{dong2021collaborative}, XR devices are able to offload the complex rendering tasks to the remote edge servers and receive high-quality scenes transmitted by wireless radio networks. 

\IEEEpubidadjcol

Adaptive bit rate (ABR) methods are extensively used by video content providers such as YouTube to deal with the dynamics of networks, thereby optimizing user experience \cite{akhtar2018oboe}. Since in the edge-enabled social metaverse streaming, the social metaverse scenes are streamed back to the users in the form of video sequences, ABR methods can also be employed to improve quality of experience (QoE). However, the social metaverse is highly interactive and has a higher requirement for latency (10 ms) and bandwidth (10 Gb/s) compared to traditional videos. The majority of the existing ABR methods which are designed for streaming non-interactive videos are not applicable to our problem. Designing an ABR method for social metaverse streaming can be very challenging due to the following reasons.

\begin{itemize}
\item User experience in the social metaverse are impacted by multiple factors, such as frame rate and latency, and these factor influence each other, making QoE hard to quantify. For example, high frame rate can make scenes look smoother, more realistic, and more immersive. But with more data transmitted by the networks, it also brings high latency which may cause sickness with XR devices \cite{fernandes2016combating}.
\item In social metaverse streaming, multiple users usually share a bottleneck network with limited bandwidth on the edge server side. The network conditions are difficult to predict with dynamic bandwidth, jitter, and delay which need to be considered when determining the appropriate bit rate. Moreover, how to allocate network resources based on users' requests is important for improving overall user experience under varying network conditions.
\item Through a trial-and-error searching strategy, deep reinforcement learning (DRL) provides a promising method to enable adaptive social metaverse streaming. It learns the policy dynamically by interacting with the environment without any pre-programmed rules. However, it is necessary to consider how to design a DRL framework that can accommodate multiple users and how to make the DRL model for multiple users converge quickly during training.
\item  Most of conventional DRL methods are operated in a centralized way where a DRL server collects data from end-users’ devices and trains data with full access. However, raw data generated by users in the social metaverse is personal and proprietary. Sharing this data to a central server can reveal sensitive information and impact industry competition. To solve this problem, we need to figure out how to collaboratively train a model with data from multiple users without any raw data leaving their devices. 
\end{itemize}

\begin{figure}[htbp]
\centering
\includegraphics[width=0.43\textwidth]{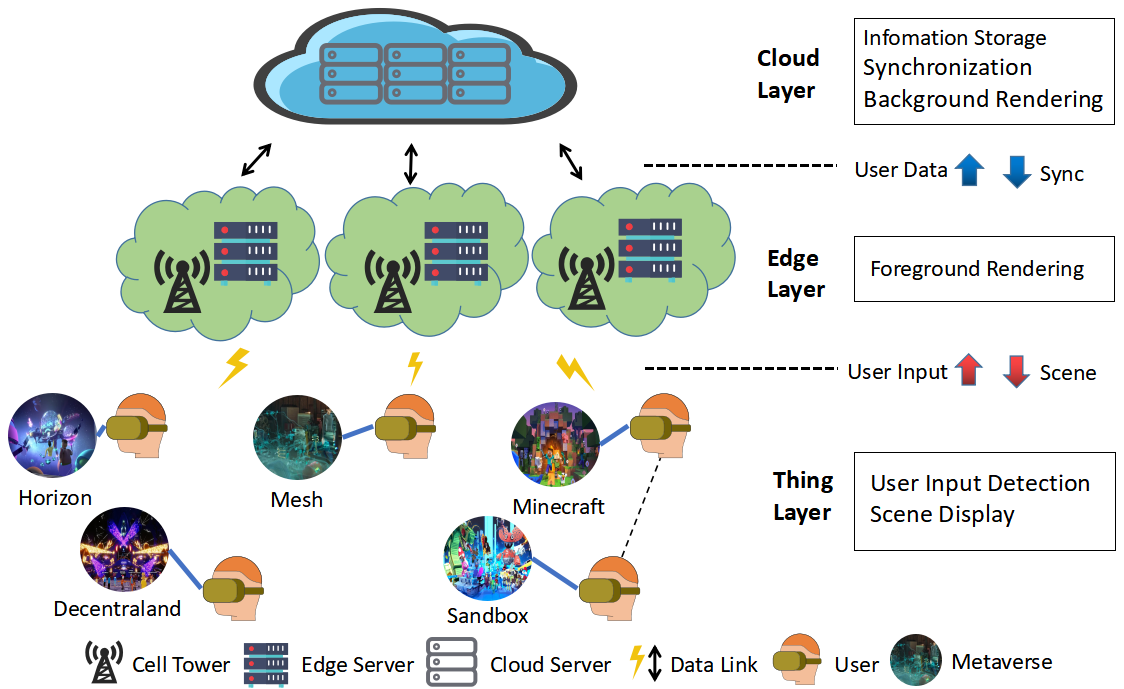}
\caption{The MEC-based architecture has three layers: the thing layer, the edge layer, and the cloud layer. These three layers collaboratively maintain seamless and efficient performance of the social metaverse in a distributed computing environment.} 
\label{mec}
\end{figure}

To solve the above challenges, we propose an MEC-based architecture of the social metaverse (shown in Fig. \ref{mec}). Users are able to upload their input data to the edge server and receive and display metaverse scenes transmitted by the cell tower. Although edge servers are designed to process data swiftly, they still rely on the cloud layer in the data center to manage computation-intensive and latency-tolerant operations. These operations include user information storage, user state synchronization, and background rendering. We then investigate the important factors related to user experience and introduce a time-step based QoE model for social metaverse streaming. Based on that, we propose federated multi-agent proximal policy optimization (F-MAPPO) which combines the advantages of federated learning (FL) and multi-agent deep reinforcement learning (MADRL) \cite{schulman2017proximal}. Specifically, to protect the privacy and security of data, F-MAPPO employs a distributed FL-based framework where multiple XR devices collaboratively learn a shared model while keeping all the training data on device. In our proposed architecture, the edge server maintains a global agent as a coordinator and each headset has a local agent as a participant. The uploading of local models and the broadcasting of global model are also delivered by the cell tower (More details will be given in Section \ref{problem}). Moreover, empowered by DRL, agents of F-MAPPO can interact with the dynamic networks and adaptively choose appropriate bit rates under varying network conditions to maximize overall QoE. In summary, the main contributions of this paper are as follows.

\begin{itemize}  
\item We design and develop ASMS, an adaptive social metaverse streaming system powered by edge computing. ASMS enables multiple users to experience high-quality, remotely rendered metaverse scenes. Edge servers handle computationally intensive rendering tasks, then stream high-resolution scenes back to users. This approach allows immersive metaverse experiences on resource-constrained headsets. Experimental results show that ASMS improves user experience metrics by at least 14\% compared to existing streaming methods.

\item We propose a time-step-based QoE model tailored for social metaverse streaming. This model incorporates key network parameters while also considering critical user experience factors, including motion-to-photon (MTP) latency, sudden network disruptions, and fluctuating user densities. To validate its effectiveness, we conducted Mean Opinion Score (MOS) evaluations, confirming its reliability in assessing user-perceived quality.

\item We formulate social metaverse streaming as a decentralized partially observable Markov decision process (Dec-POMDP) and introduce F-MAPPO, which enables multiple agents to dynamically optimize bit rate selection in unpredictable network conditions. Unlike existing learning-based methods that overlook privacy concerns, ASMS integrates local differential privacy (LDP) to protect user data during training. We also evaluate its computational and communication overhead, showing that ASMS is practical and feasible for real-world deployment without excessive resource consumption.

\end{itemize}

The remainder of this paper is organized as follows. Section \ref{related} provides the background of this work. The problem formulation and method design of F-MAPPO is described in Section \ref{problem}. The evaluation results are summarized in Section \ref{experiment} followed by a conclusion in Section \ref{conclusion}.  

\section{Background} \label{related}
% In this section, we begin with a concise overview of the various stages of the social metaverse, outlining the features and associated technologies at each stage from its inception to the present. Subsequently, we review the issue of data privacy within the social metaverse and the integration of FL methods with other technologies to address this challenge. Finally, we delve into the latest advancements in adaptive bit rate control methods for video streaming.
\subsection{Development of the Social Metaverse}
The development of the social metaverse can be envisioned through several stages, each representing a significant evolution in technology, infrastructure, and user engagement. 
\subsubsection{Early Concepts (1960s-2000s)} 
The initial conceptualization of the social metaverse was largely inspired by science fiction and nascent virtual environments, which served as a catalyst for the first iterations of online social interaction. This foundational period witnessed the emergence of early VR experiments and the establishment of basic Internet infrastructure, setting the stage for subsequent technological advancements. Seminal works, notably Neal Stephenson's ``Snow Crash," along with pioneering platforms such as ``Second Life" \cite{kaplan2009fairyland} and ``The Sims Online," \cite{martey2007digital} epitomize the innovative spirit and exploratory nature of this formative era.
\subsubsection{Technological Advancements (2000s-2015s)}
This period was marked by significant advancements in connectivity and the emergence of more sophisticated and immersive virtual environments. Key developments included the widespread adoption of broadband Internet, the advent of cloud computing, and the introduction of advanced VR headsets such as the Oculus Rift and HTC Vive, alongside AR devices like Microsoft HoloLens. Massively multiplayer online games, exemplified by ``World of Warcraft," \cite{chen2009communication} and the growing utilization of VR in gaming, education, and training further defined this era, reflecting the expanding scope and potential of VR technologies.
\subsubsection{Virtual Economies (2015s-2020s)}
This stage is characterized by the seamless integration of cross-platform experiences and the creation of complex virtual economies and digital assets. During this period, there was a significant rise in unified development platforms and the widespread adoption of blockchain technology, leading to the emergence of Non-Fungible Tokens (NFTs) \cite{ali2023review}. This era is exemplified by games such as ``Fortnite" \cite{carter2020situating} and ``Minecraft VR," as well as platforms like ``Roblox VR," which support cross-platform play and decentralized virtual asset marketplaces, showcasing the advancements in virtual environment interconnectivity and economic innovation.
\subsubsection{Integrated Metaverse (2020s-present)}
The current stage involves the creation of continuous, interconnected virtual spaces and the widespread integration of the social metaverse into daily life, encompassing work, education, and social interaction. Key advancements in VR/AR technologies, robust online infrastructure, AI-driven environments, and decentralized applications are central to this phase. Platforms such as Meta’s Horizon Worlds, Microsoft’s Mesh, and blockchain-based virtual worlds like Decentraland \cite{guidi2022social} exemplify this era, highlighting the merging of digital and physical realities and the expansion of virtual environments into mainstream activities.
\subsection{FL for the Social Metaverse}
FL was developed to tackle the task of safeguarding user privacy while simultaneously harnessing the potential of extensive data for training machine learning models \cite{mcmahan2017communication}. When combined with metaverse technologies like communication technology, MEC, and blockchain, FL becomes a potent solution for addressing privacy concerns within the social metaverse and enhancing resource sharing \cite{chen2023federated}. For instance, Yang et al. proposed FL's application in 6G networks to facilitate high-performance data communication within the social metaverse, conserving wireless resources and reducing transmission latency \cite{yang2022federated}. Model parameters are transmitted from individual terminal devices to the server in a coordinated manner. Incorporating FL into MEC technology ensures that users' sample data remains securely stored on their respective devices \cite{yu2021toward}, mitigating risks associated with attacks or failures among a small number of edge devices. In this way, the social metaverse avoids significant data privacy exposure or service disruptions. Concerning blockchain, a novel approach \cite{kim2019blockchained} replaces the central aggregator with a peer-to-peer blockchain network, utilizing blockchain nodes for FL model aggregation. This distributed storage mechanism guarantees user identity authentication and safeguards the security of digital assets within the social metaverse.

\subsection{Learning-Based Adaptive Bit Rate Control}
In ABR streaming, the bit rate of the encoding is dynamically adjusted to ensure high-quality video delivery across a wide range of network conditions \cite{rainer2016statistically}. Inspired by the success of DRL methods in solving sequential decision-making problems, some recent works have been focusing on solving the ABR streaming problem with DRL. Mao et al. proposed Pensieve, a system that learns ABR methods automatically by the popular DRL Actor-Critic framework \cite{mao2017neural}. Huang et al. proposed Tiyuntsong which is a self-play DRL approach with generative adversarial network (GAN)-based method for ABR video streaming \cite{huang2019tiyuntsong}. As both social metaverse streaming and cloud/edge gaming are latency sensitive and require massive computing and communication resources, there are also some DRL-based ABR methods for cloud/edge gaming. Chen et al. presented a method for dynamically encoding video frames in cloud gaming based on WebRTC, whereby the default Google Congestion Control (GCC) is replaced by a DRL method \cite{chen2019t}. The neural networks are trained by A3C \cite{babaeizadeh2016reinforcement}, another widely used DRL method to maximize the QoE value.

Despite these advancements, existing ABR streaming solutions face significant limitations in the metaverse. First, the metaverse requires much stricter video streaming conditions than traditional applications like video-on-demand or cloud gaming. It demands ultra-low latency, ultra-high bandwidth, and real-time adaptability to support immersive, interactive experiences. Most existing ABR solutions, designed for conventional streaming, struggle to meet these performance demands. Second, previous DRL-based ABR approaches focus on optimizing streaming for a single user, while the metaverse typically involves multiple users sharing network resources in dynamic environments. Existing methods lack effective coordination for bit rate allocation across multiple users, leading to inefficient resource utilization and degraded user experience. Third, privacy concerns remain largely unaddressed in current DRL-based ABR models, which often rely on centralized training that requires access to raw user data. In the metaverse, where sensitive biometric and behavioral data are continuously collected, this raises serious privacy risks. To overcome these challenges, we propose F-MAPPO, which integrates MADRL with FL. This approach allows multiple users to collaboratively optimize streaming quality without sharing raw data. By keeping sensitive information on local devices, F-MAPPO enhances privacy, improves network resource allocation, and adapts efficiently to dynamic conditions.

\section{Problem Formulation and Method Design} \label{problem}
To achieve adaptive, high-quality streaming in the social metaverse, we model the problem as a multi-agent decision-making task. Multiple users interact and compete for shared network resources while experiencing immersive virtual environments. This section first defines the problem within a MADRL framework, incorporating a novel QoE model to measure user satisfaction in the metaverse. Since metaverse interactions involve sensitive user data, FL is integrated to enable decentralized model training. This allows user devices to collaboratively learn an optimal streaming policy while keeping raw data local, preserving privacy without compromising performance.

\subsection{Agent Design}
The definitions of the state space, the action space, and the reward for F-MAPPO are introduced.

\begin{itemize}
\item State: The state of an agent at time step $t$ is defined as: $s_t = (x_t, y_t, l_t, j_t, p_t, n_t)$, where $x_t$ is the last selected target bit rate; $y_t$ is the actually received bit rate;   $l_t$ is the average MTP latency; $j_t$ is the network jitter which denotes the mean variation in RTT; $p_t$ is the number of lost packets; and $n_t$ is the number of sent negative acknowledgment messages. At every time step, these six parameters are calculated and they are considered to be representative of the general performance of the local networks \cite{winstein2013tcp}. 
\item Action: An action of an agent at time step $t$ $a_t$ specifies how much change should be made to the last selected bit rate based on the local observation. By specifying a positive, negative, or 0 value, the bit rate will be increased, reduced, or remain the same. 
\item Reward: After applying the joint bit rate adjustment for each agent, the environment returns a global reward $r_t$ for that time step. It is important that the global reward is able to reflect the overall consequence of the joint action. Using a predefined QoE model for social metaverse streaming, we can quantify the local rewards for each agent, and then average the QoE values to obtain the global reward.
\end{itemize}

% \begin{table}[htbp]
% \centering
% \caption{NOTATIONS AND EXPLANATIONS}
% \label{paras}
% \begin{adjustbox}{width=0.46\textwidth}
% \begin{tabular} {|c|c|}
% \hline
% \textbf{Notation} & \textbf{Explanation}\\
% \hline
% $x_t$ & last selected target bit rate at time step t\\
% \hline
% $y_t$ & actually received bit rate at time step t\\
% \hline
% $l_t$ & average round trip latency at time step t\\
% \hline
% $j_t$ & network jitter at time step t\\
% \hline
% $p_t$ & number of lost packets at time step t\\
% \hline
% $n_t$ & number of sent negative acknowledgment message at time step t\\
% \hline
% \end{tabular}
% \end{adjustbox} 
% \end{table}

Currently, most of the QoE models available for social metaverse systems, such as VR headsets, are designed to evaluate the performance of local rendering performed on the XR devices. There is a strong correlation between the computation resources on these devices and the factors contributing to the QoE model, such as FPS and resolution \cite{long2022interacting}. Social metaverse streaming, on the other hand, provides users with more computation resource from edge servers, but they are highly sensitive to the extra latency introduced by encoding, network delay, and decoding. These steps are necessary for streaming and displaying the social metaverse scene frames. Since the performance of social metaverse streaming changes dynamically at each time step, we propose a time-step-based QoE model. This model integrates key network parameters and user experience factors, including MTP, sudden network disruptions, and fluctuating user densities. It is defined as follows:

\begin{equation}
\label{qoe}
\begin{split}
QoE_t = & \underbrace{\alpha q(y_t) \cdot e^{- \frac{u_t}{u_{\text{max}}}}}_{\text{Scene Quality with User Density Impact}} 
- \underbrace{\beta |y_t - f_{target}|}_{\text{Choppiness Penalty}} 
\\
& - \underbrace{\gamma \cdot \frac{l_t}{y_t + \epsilon}}_{\text{MTP Latency Penalty}}  - \underbrace{\delta_1 |q(y_{t+1}) - q(y_t)| + \delta_2 \cdot P(p_t)}_{\text{Stability and Disruption Penalty}}
\end{split}
\end{equation}

For the time step \( t \), \( \alpha q(y_t) \cdot e^{- \frac{u_t}{u_{\text{max}}}} \) represents the overall satisfaction with the scene quality, where \( y_t \) is the average received bit rate, \( q(y_t) = \log  (y_t / y_{\min}) \) is a logarithmic function capturing the diminishing returns of increasing bit rate, \( u_t \) is the current user count, and \( u_{\text{max}} \) is the maximum user capacity. The higher the bit rate and the lower the user density, the better the scene quality and the more enjoyable the viewing and interaction experience in the social metaverse.

The choppiness penalty, \( \beta |f_t - f_{target}| \), quantifies the negative impact of mismatched frame rate \( f_t \) relative to the target frame rate \( f_{target} \), with \( \beta \) being the penalty factor. The latency impact, \( \gamma \cdot l_t / (y_t + \epsilon) \), accounts for the influence of MTP latency \( l_t \) on QoE, adjusted by the received bit rate \( y_t \) to reflect the mitigative effects of higher throughput. Here, \( \epsilon \) is a small constant \( (10^{-6}) \) to prevent division by zero.

The stability and disruption penalty has two components: \( \delta_1 |q(y_{t+1}) - q(y_t)| \) measures the penalty for scene quality fluctuations between consecutive time steps \( t \) and \( t+1 \), while \( \delta_2 P(p_t) \) penalizes packet losses. The term \( P(p_t) = \max(0, p_t - p_{\text{threshold}}) \) is introduced specifically to address sudden network disruptions, where \( p_t \) is the number of lost packets and \( p_{\text{threshold}} \) is a predefined threshold. Packet loss is penalized only when it exceeds this threshold. This prevents minor network fluctuations from disproportionately affecting QoE while emphasizing stability during major disruptions.

This QoE model integrates scene quality, MTP latency, user density, and stability factors, ensuring a comprehensive evaluation of user experience in dynamic network environments typical of the social metaverse.

\subsection{Local Model Training of Agents}

After modeling the state space, observation space, the action space, and the global reward function of social metaverse streaming, a MADRL method is employed to improve the average QoE in our proposed system. In MADRL, each agent interacts with the environment according to a policy, which is a rule that specifies how the action should be carried out. When we propose a neural network to approximate a stochastic policy function, the output of the network depends on the weights and biases of the neural network parameterized by $\theta$. The objective of the agent is to find the optimal policy $\pi_{\theta}$ that maximizes the accumulated reward:
\begin{equation}
G_t = \sum^{\infty}_{k=0} \gamma^kr_{t+k+1}
\end{equation}
where $\gamma$ is the discount coefficient that trades off the weights of historical and current rewards for the policy. The state value function under a policy $\pi_{\theta}$ is represented by:
\begin{equation}
V^{\pi_{\theta}}({s_t}) = \mathbb {E}_{\pi_{\theta}} [G_t|{s}_t = {s}] = \mathbb {E}_{\pi_{\theta}} \bigg [ \sum^{\infty}_{k=0} \gamma^kr_{t+k+1}|{s}_t = {s} \bigg ]
\end{equation}

To better evaluate decision-making under uncertainty, we define the state-action value function $Q^{\pi_{\theta}}(s_t, a_t)$. This function measures the expected cumulative reward when taking action $a_t$ in state $s_t$, followed by subsequent actions determined by policy $\pi_{\theta}$. It allows the agent to assess the long-term value of each action in a given state:  
\begin{equation}
\begin{split}
Q^{\pi_{\theta}}({s_t}, {a_t}) & = \mathbb {E}_{\pi_{\theta}} [G_t|{s}_t = {s}, {a}_t = {a}] \\ & = \mathbb {E}_{\pi_{\theta}} \bigg [ \sum^{\infty}_{k=0} \gamma^kr_{t+k+1}|{s}_t = {s}, {a}_t = {a} \bigg ]
\end{split}
\end{equation}

While the above Q-function provides an absolute measure of expected rewards, it is often useful to compare an action’s value relative to the policy's average performance. This motivates the advantage function, which quantifies how much better or worse taking a specific action $a_t$ at state $s_t$ is compared to the expected value of following policy $\pi_{\theta}$. The advantage function helps reduce variance in policy updates and improves learning efficiency:  
\begin{equation}
A^{\pi_{\theta}}({s}_t, {a}_t) = Q^{\pi_{\theta}}({s}_t, {a}_t) - V^{\pi_{\theta}}({s}_t)
\end{equation}

The goal of DRL is to maximize the agent’s long-term expected reward. This is expressed through the policy objective function, which represents the expected value of all states under the policy $\pi_{\theta}$. The policy optimization process seeks to adjust $\theta$ to maximize this function, ensuring that the agent consistently selects actions that yield higher rewards over time:  
\begin{equation}
\begin{split}
J(\theta) & = \sum_{s_t \in \mathcal{S}}d^{\pi_{\theta}}({s}_t)V^{\pi_{\theta}}({s}_t) \\ & = \sum_{{s}_t \in \mathcal{S}}d^{\pi_{\theta}}({s}_t)\sum_{a_t \in \mathcal{A}} Q^{\pi_{\theta}}({s}_t, {a}_t)\pi_{\theta}({a}_t|{s}_t)
\end{split}
\end{equation}
where $d^{\pi_{\theta}}({s}_t)$ is the stationary distribution of the Markov chain for $\pi_{\theta}$.

To maximize the above objective value, the gradient of $J({\theta})$ concerning ${\theta}$ is calculated as:
\begin{equation}
\begin{split}
\nabla_{{\theta}}J({\theta})& = \nabla_{{\theta}}\sum_{s_t \in \mathcal{S}}d^{\pi_{\theta}}({s}_t)\sum_{{a}_t \in \mathcal{A}} Q^{\pi_{\theta}}({s}_t, {a}_t)\pi_{\theta}({a}_t|{s}_t)\\ &\propto \sum_{s_t \in \mathcal{S}}d^{\pi_{\theta}}({s}_t)\sum_{{a}_t \in \mathcal{A}} Q^{\pi_{\theta}}({s}_t, {a}_t)\nabla_{{\theta}}\pi_{\theta}({a}_t|{s}_t)\\ & = \sum_{s_t \in \mathcal{S}}d^{\pi_{\theta}}({s}_t)\sum_{{a}_t \in \mathcal{A}} \pi_{\theta}({a}_t|{s}_t) Q^{\pi_{\theta}}({s}_t, {a}_t)\frac{\nabla_{{\theta}}\pi_{\theta}({a}_t|{s}_t)}{\pi_{\theta}({a}_t|{s}_t)}\\ & = \mathbb {E}_\pi \bigg [Q^{\pi_{\theta}}({s}_t, {a}_t)\nabla_{{\theta}}log\pi_{\theta}({a}_t|{s}_t) \bigg ] \\ & = \mathbb {E}_\pi \bigg [A^{\pi_{\theta}}({s}_t, {a}_t)\nabla_{{\theta}}log\pi_{\theta}({a}_t|{s}_t) \bigg ] \\
\end{split}
\end{equation}

where $\mathbb {E}_\pi$ refers to $\mathbb {E}_{{s}\sim d_{\pi_{\theta}}, {a}\sim{\pi_{\theta}}}$ when both state and action distributions follow the policy $\pi_{\theta}$. Therefore, gradient ascent can be applied to update the parameters of the policy. Proximal Policy Optimization (PPO), a popular model-free on-policy DRL method, has shown significant effectiveness in selecting appropriate step sizes in single-agent environment by limiting the change of the policy at each step. During the data collection period, each agent interacts with the environment with the old policy network $\pi_{\theta_{old}}$. The collected trajectory with $T$ time steps can be denoted as $\tau = \{{s}_0, {a}_0, r_0,...,{s}_{T-1}, {a}_{T-1}, r_{T-1}, {s}_t\}$. In the training process, PPO provides a clipped surrogate objective denoted as:
\begin{equation}
J^{CLIP}(\theta) = \mathbb {E}_{{s}_t, {a}_t} \bigg [min \bigg (r(\theta) \hat A({s}_t, {a}_t), g(\epsilon, \hat A({s}_t, {a}_t))\bigg ) \bigg ]
\end{equation}
where 
\begin{equation}
g(\epsilon, \hat A({s}_t, {a}_t)) = \left \{
                      \begin{array}{ll}
                          (1 + \epsilon) \hat{A}({s}_t, {a}_t),\ \hat{A}({s}_t, {a}_t) \geq 0  \\
                          (1 - \epsilon) \hat{A}({s}_t, {a}_t),\ \hat{A}({s}_t, {a}_t) < 0
                      \end{array}
                      \right.
\end{equation}
and $r(\theta) = \pi_{\theta}(a_t|s_t)/\pi_{\theta_{old}}(a_t|s_t)$ is the ratio between the new policy and the old one and $\epsilon$ is a hyperparameter (usually 0.1 or 0.2). According to the above functions, the probability ratio $r(\theta)$ is limited within $[1 - \epsilon, 1 + \epsilon]$ when the estimated advantage function $\hat A({s}_t, {a}_t)$ is positive or negative so that the policy can only change in a small range in both cases. The advantage estimate $\hat A({s}_t, {a}_t)$ is calculated based on the collected trajectory $\tau$ using the generalized advantage estimation (GAE) \cite{schulman2015high}:
\begin{equation}
\hat{A}({s}_t, {a}_t) = \sum^{T-1-t}_{l=0}(\gamma \lambda)^{l}\delta_{t+l}
\label{gae}
\end{equation}

where $\delta_t$ is the temporal difference error denoted as $\delta_t = r_t + \gamma V^{\pi_{\theta}}({s}_{t+1}) - V^{\pi_{\theta}}({s}_t)$ and $\lambda \in [0, 1]$ is a hyperparameter to balance the bias-variance trade-off. The estimated discounted return $\hat{G}_t$ based on the collected trajectory $\tau$ is computed by the truncated GAE as follows:
\begin{equation}
\hat{G}_t = \sum^{T-t}_{i=0}\gamma^{i}r_{t+i} + \gamma^{T-t}V^{\pi_{\theta}}({s}_t)
\label{edr}
\end{equation}
where $\gamma \in (0, 1]$ is the discount factor and $V^{\pi_{\theta}}$ is the global value function. For a set of trajectories $\mathcal{D} = \{\tau\}$, the parameters of the policy network in an agent are updated by maximizing the average value of the clipped surrogate objective for all steps:
\begin{equation}
\theta = \argmax_{\theta}\frac{1}{|\mathcal{D}|T}\sum_{\tau \in \mathcal{D}}\sum^{T-1}_{t=0}J^{CLIP}(\theta)
\label{ascent}
\end{equation}
The value network is updated by minimizing the average value of the mean-squared error between the estimated returns and the calculated returns by the value function:
\begin{equation}
{w} = \argmin_{{w}}\frac{1}{|\mathcal{D}|T}\sum_{\tau \in \mathcal{D}}\sum^{T-1}_{t=0}(V^{\pi_{\theta}}({s}_t) - \hat{G}_t)^2
\label{descent}
\end{equation}

\subsection{FL Design}

\begin{figure}[htbp]
\centering
\includegraphics[width=0.98\linewidth]{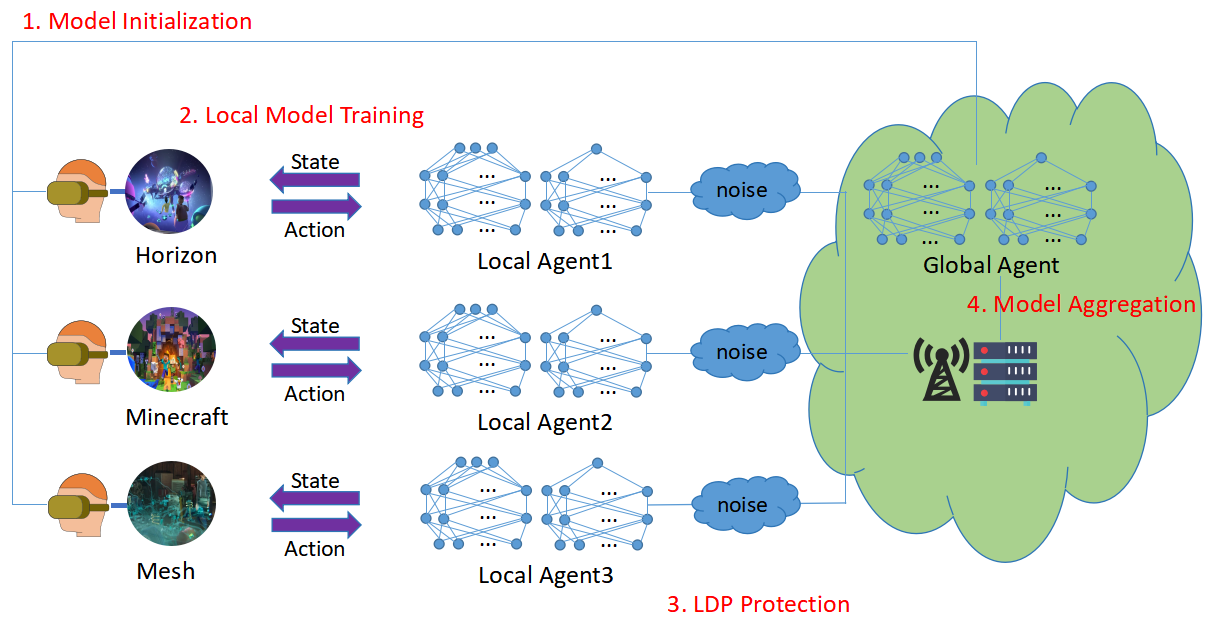}
\caption{ The FL training process of F-MAPPO with LDP for MEC-based social metaverse streaming.} 
\label{fl}
\end{figure}

As elaborated earlier, the deployment of multiple agents on individual local headsets serves to regulate bit rates for respective users. However, this decentralized configuration raises concerns regarding generalization performance. The confined diversity of data within isolated devices heightens the risk of agents becoming ensnared within local optima. To address this issue, we introduce F-MAPPO, which leverages FL to improve data generalization during training. At the same time, it safeguards the security and privacy of user-generated data in the social metaverse. To further enhance privacy protection, LDP is integrated into the FL process. This ensures that sensitive user data is protected before being shared with the global model. The main steps of F-MAPPO are outlined below (shown in Fig. \ref{fl}).

\begin{itemize}
\item Step 1: Model Initialization: 
In the process of setting up a client-server-based learning system, the global agent creates an initial model and sends it to each local agent, which means all the agents have the same actor-critic structure and parameters of the network.

\item Step 2: Local Model Training:  
After initializing the network model, each agent starts to collect its own dataset by interacting with the environment. Specifically, each agent executes an action about how to adjust the bit rates of streaming frames depending on the observed network condition. It then receives the global reward from the environment represented by:  \( r = r_1 + r_2 +...+ r_I\). Each agent trains a local model to maximize the global reward based on its respective dataset.

\item Step 3: LDP Protection: 
Before transmitting the local model updates, LDP is applied to ensure privacy protection. Each agent perturbs its local model gradients using the Laplacian mechanism:

\begin{equation}
\tilde{\theta} = \theta + \text{Lap}\left(\frac{\Delta \theta}{\epsilon}\right)
\label{LDP}
\end{equation}

where $\tilde{\theta}$ represents the perturbed model update, $\epsilon$ is the privacy budget, and $\Delta \theta$ denotes the sensitivity of gradient updates. This process ensures that no raw model parameters are exposed, mitigating privacy risks.

\item Step 4: Model Aggregation:  
When an episode is finished, each headset transmits the perturbed local model update $\tilde{\theta}$ instead of raw gradients to the central server. The central server then updates and trains the new global model for the next episode using the Federated Averaging (FedAvg) method \cite{mcmahan2017communication}, with additional weighted aggregation to mitigate the impact of LDP noise and data heterogeneity among users:

\begin{equation}
\theta^{G} = \sum_{i=1}^{N} \frac{w_i}{\sum_{j=1}^{N} w_j} \tilde{\theta}_i
\label{FedAvg}
\end{equation}

where $w_i$ is dynamically adjusted.

\item Step 5: Iterative Learning Process:  
Repeat the above-mentioned process until the global reward achieves the target value or the maximum number of iterations is reached.

\end{itemize}

The complete training steps for each agent in F-MAPPO are detailed in Algorithm 1. At the start of each episode, the agent updates its policy and value networks using the latest global model parameters received from the FL server. During the episode, the agent interacts with the environment, selects actions based on its current policy, and collects rewards. Once it accumulates enough experience, it computes advantage estimates and discounted returns, storing them in a memory buffer for future optimization. The policy and value networks are then trained using mini-batch updates via gradient ascent and descent. Finally, the updated parameters are perturbed using LDP before being sent back to the FL server for global aggregation.

% For the convenience of readers, the notations used in the paper are summarized in Table 1.

\begin{algorithm} 
    \renewcommand{\algorithmicrequire}{\textbf{Input:}}
	\renewcommand{\algorithmicensure}{\textbf{Output:}}
	\caption{Training Procedure for Agents in F-MAPPO} 
	\label{alg1} 
	\begin{algorithmic}[1]
		\STATE Initialize value network and policy network with parameters ${w}$ and ${\theta}$ respectively 
		\STATE Initialize target value network with $\overline {w} \gets w$ and old policy network with $\theta_{old} \gets \theta$ 
		\FOR{episode $e = 1, 2, ...$}
            \STATE Update value network and policy network with parameters from the FL server $\{w^G_{e-1}, \theta^G_{e-1} \}$
		    \FOR{time step $t = 1, 2, ..., T$}
                \STATE Execute an action according to $\pi_{\theta_{old}}(a_t|o_t)$
                \STATE Get the global reward $r_t$, and the next environment state ${s}_{t+1}$
            \ENDFOR
    		\FOR{epoch k = $1, 2, ...$}
    		    \STATE Get a set of trajectories $\mathcal{D}_k = \{\tau\}$
    		    \STATE Compute the estimated advantages $\hat A({s}_t, {a}_t)$ according to Equation \ref{gae} 
    		    \STATE Compute the estimated discounted return $\hat{G_t}$ using Equation \ref{edr} 
                \STATE Initialize a memory buffer $M_k$
                \STATE Store data $\{o_t, a_t, \hat A({s}_t, {a}_t), \hat G_t\}^{|\mathcal{D}_k|T}_{t=1}$ into $M_k$
                \STATE Shuffle and reorder the data in $M_k$
                \STATE Select groups of data from $M_k$:
                \STATE Apply mini-batch gradient ascent on $\theta$ using Equation  \ref{ascent} 
                \STATE Apply mini-batch gradient descent on ${w}$ using Equation  \ref{descent}
    		\ENDFOR
            \STATE Upload the parameters of value network and policy network $\{w_{e}, \theta_{e} \}$ to the FL server
    	\ENDFOR
	\end{algorithmic} 
\end{algorithm}

\section{Experiment and Analysis} \label{experiment}
% In this section, we present a case study on the adaptive allocation of communication and computation resources for users in the ASMS. We provide a detailed description of the architecture of our intelligent resource allocation system and outline the three evaluation baselines employed in our study. Finally, we demonstrate the analysis of the results from offline training and online testing, highlighting the system's effectiveness and performance. 

\begin{figure*}[htbp]
\centering
\includegraphics[width=0.95\textwidth]{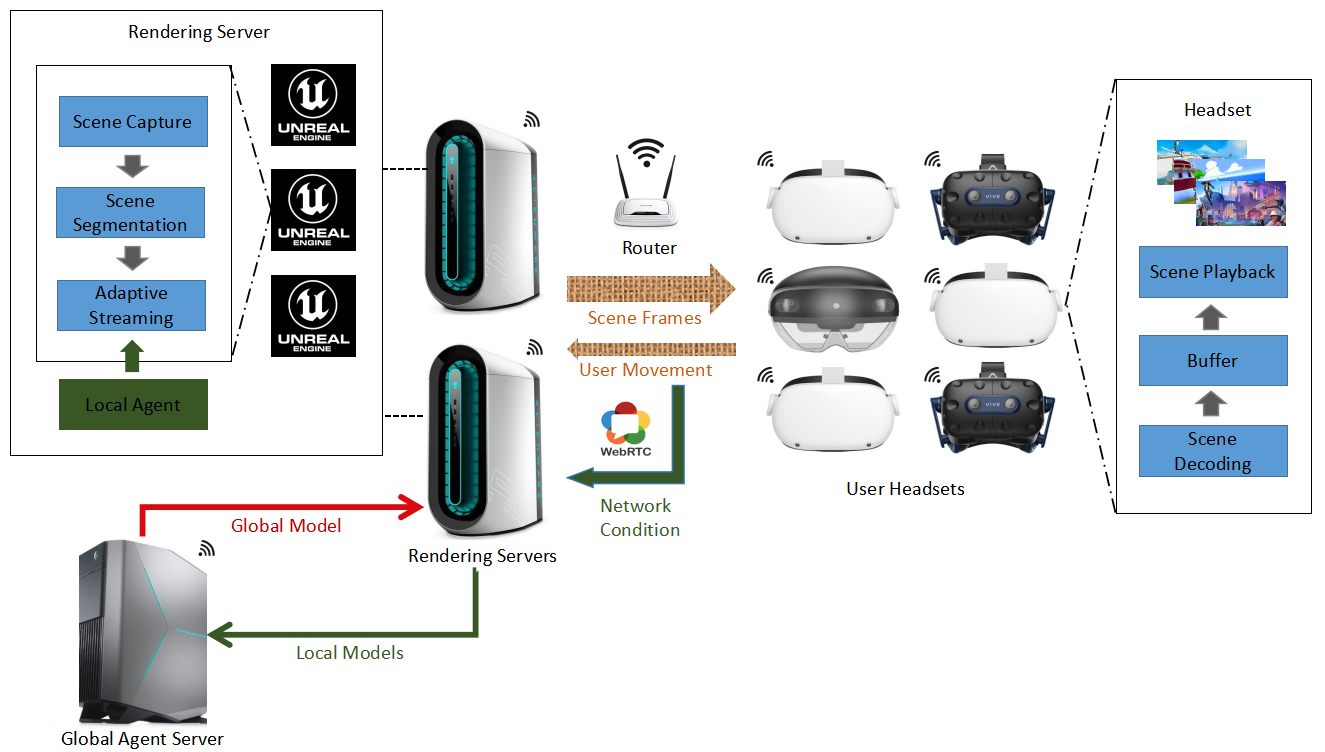}
\caption{  The ASMS architecture consists of two rendering servers, a global agent server, and multiple VR/MR headsets, all connected via a wireless router. The rendering servers handle social metaverse scene capture, segmentation, encoding, and adaptive streaming of scene frames to the headsets. The global agent server aggregates policy updates and coordinates learning across devices. Once received, the headsets decode, buffer, and play the frames, allowing users to experience high-quality, immersive metaverse environments in real time.} 
\label{system}
\end{figure*}

\subsection{Experimental Setup}
In our experiments (illustrated in Fig. \ref{system}), we utilize two rendering servers, each equipped with the Windows 10 operating system, an Intel Core i9-11900F processor, 64 GB of RAM, and an NVIDIA GeForce RTX 3090 graphics card. The social metaverse scenarios are created using Unreal Engine 5.3\footnote{https://www.unrealengine.com/en-US}, allowing real-time streaming of metaverse scenes from the rendering servers to external users via WebRTC. A global agent server, equipped with an NVIDIA GeForce RTX 4090 GPU, coordinates the FL process during the offline training. The user-side devices include three Meta Quest 2 headsets, two HTC Vive headsets, and one HoloLens 2, providing a diverse range of metaverse experiences.

To transmit scene frames and movement information between the rendering servers and users, all devices connect to a single TP-Link router via wireless links. Communication between the local agents, the global agent, and the Unreal Engine instances is managed through Socket.IO\footnote{https://python-socketio.readthedocs.io/en/latest/}. To simulate diverse network conditions, we utilize Clumsy\footnote{https://github.com/jagt/clumsy} to introduce variable delays, bandwidth constraints, and random packet loss rates.

\subsection{Parameter Optimization of QoE Model}

To determine the penalty coefficients (\( \alpha, \beta, \gamma, \delta_1, \delta_2 \)) in the proposed QoE model (Equation  \ref{qoe}) and validate its robustness, we conducted a user study involving 8 participants. We first simulated 6 typical scenarios in the social metaverse (shown in Table. \ref{network_conditions}) including high-performance cloud streaming (s1), home Wi-Fi 6 network (s2), 4G LTE mobile network (s3), 5G edge computing (s4), network congestion (s5), and network recovery (s6). Each participant experienced a scenario for 5 minutes, during which participants rated their QoE using a 5-point MOS scale (1 = very poor, 5 = excellent). Each participant repeated the experiment 4 times to ensure consistency and reliability of the results. Overall, each participant evaluated 24 scenarios, resulting in 192 total MOS ratings across all participants. The scores were averaged for each scenario to reduce individual variability and ensure a robust dataset for analysis. 

\begin{table*}[htbp]

\centering
\caption{Typical Network Conditions for Social Metaverse Streaming}
\label{network_conditions}
\begin{adjustbox}{width=0.95\textwidth}
\begin{tabular} {|c|c|c|c|c|c|}
\hline
\textbf{Scenario} & \textbf{Bandwidth (Mbps)} & \textbf{Latency (ms)} & \textbf{Jitter (ms)} & \textbf{Packet Loss (\%)} & \textbf{Burst Loss (\%)}\\
\hline
High-performance cloud streaming (S1) & 100 - 200 & 10 - 30 & 2 - 5 & 0.1 & 0 \\
\hline
Home Wi-Fi 6 network (S2) & 50 - 100 & 30 - 50 & 5 - 10 & 0.5 & 0.5 \\
\hline
4G LTE mobile network (S3) & 20 - 80 & 50 - 100 & 10 - 20 & 1 & 1 \\
\hline
5G  edge computing (S4) & 200 - 500 & 5 - 10 & 1 - 3 & 0.1 & 0 \\
\hline
Network congestion (S5) & 100 $\rightarrow$ 30 & 50 $\rightarrow$ 100 & 5 $\rightarrow$ 20 & 0.5 $\rightarrow$ 5 & 10 \\
\hline
Network recovery (S6) & 30 $\rightarrow$ 100 & 100 $\rightarrow$ 20 & 20 $\rightarrow$ 5 & 2 $\rightarrow$ 0.5 & 5 $\rightarrow$ 0\\
\hline
\end{tabular}
\end{adjustbox}
\end{table*}

A grid search optimization was performed to minimize the root mean square error (RMSE) between the model-predicted QoE and MOS ratings. This process resulted in the following coefficients: \( \alpha = 1 \), \( \beta = 0.4 \), \( \gamma = 0.2 \), \( \delta_1 = 0.6 \), and \( \delta_2 = 0.5 \). The optimized model achieved a strong correlation with MOS ratings (\( R^2 = 0.92 \)), confirming its ability to capture user-perceived QoE across diverse network scenarios. To test the robustness of the model, a sensitivity analysis was conducted by varying each coefficient by ±20\% while keeping others constant. The results showed an average RMSE change of 5.2\%, demonstrating that the model remains effective even when penalty coefficients shift due to network variations.

\subsection{Performance Evaluation Benchmarks}
To evaluate the performance of F-MAPPO for social metaverse streaming,  7 baselines are introduced:
\begin{enumerate}
\item 
Independent Proximal Policy Optimization (IPPO): is a decentralized learning approach that decomposes a MADRL problem with $n$ agents into $n$ decentralized single-agent problems. In this framework, each agent treats the other agents as part of the environment and learns its policies based on its own local observations. This method allows each agent to operate independently, focusing solely on optimizing its own actions without needing centralized coordination.
\item Soft Actor-Critic (SAC) \cite{haarnoja2018soft}: distinguishes itself from other RL methods by striving to maximize both the discounted cumulative rewards and the entropy of the policy. The inclusion of entropy maximization aims to enhance the randomness in the policy, promoting more exploratory behavior and improving overall robustness and performance. To mitigate the risk of breaching personal information during the centralized training process, we implement an independent variant of SAC. This approach ensures that each agent operates autonomously, thereby safeguarding personal data while maintaining the advantages of SAC.

\item GreenABR \cite{turkkan2022greenabr}: integrates DRL to optimize streaming quality while minimizing energy consumption. It considers perceptual video quality and real power measurements to balance bitrate selection and device energy usage, making it particularly relevant for mobile users. 

\item Ruyi \cite{10739347}: profiles user-specific QoE preferences and assigns different weights to quality metrics such as video resolution, rebuffering, and smoothness. It leverages supervised learning to predict the impact of different bitrate decisions on user satisfaction, allowing it to dynamically adjust streaming quality to maximize individual user experience.

\item ARTEMIS \cite{295533}: is a bitrate ladder optimization system designed for live video streaming. It dynamically configures bitrate ladders based on content complexity, network conditions, and client statistics. By leveraging log data of content delivery networks and real-time encoding quality indicators, ARTEMIS optimizes video streaming performance while reducing computational overhead.

\item GCC \cite{carlucci2016analysis}: is integrated into Chrome's WebRTC stack for real-time video streaming. It employs an adaptive threshold to dynamically adjust the sending rate based on estimated network delay, ensuring efficient and responsive data transmission. This method helps maintain video quality and reduce latency by continuously adapting to network conditions.
\item Bottleneck Bandwidth and Round-trip Propagation Time (BBR) \cite{cardwell2017bbr}: is a congestion control method developed by Google, designed to optimize both throughput and round-trip time (RTT). BBR achieves this by estimating the bottleneck bandwidth and RTT, and then using these estimates to compute an optimal pacing rate for data transmission. This method aims to balance high network efficiency with low latency, resulting in improved overall performance for data flow.
\end{enumerate}

\subsection{Offline Training}

The three DRL-based methods (F-MAPPO, IPPO, and SAC) use an actor-critic architecture, where each agent has an actor network and a critic network. Both networks consist of two fully connected layers. The actor network uses the tanh activation function, while the critic network applies ReLU. For stable training, mini-batch gradient updates were performed using the Adam optimizer. The policy networks were updated every 40 time steps. In F-MAPPO, the federated averaging process was conducted every four iterations, enabling collaborative learning while preserving user privacy. All key hyperparameters used in the experiments are summarized in Table \ref{paras}. To ensure fairness, all methods were configured with the same parameter settings.

\begin{table}[htbp]
\centering % 确保表格居中
\caption{Parameter Specifications in the Experiments}
\label{paras}
\resizebox{0.4\textwidth}{!}{ % 缩放表格
\begin{tabular} {|c|c|}
\hline
\textbf{Parameters} & \textbf{Value} \\
\hline
Optimizer & Adam \\
\hline
Number of neurons in hidden layers & 128 \\
\hline
Activation function (actor) & tanh \\
\hline
Activation function (critic) & ReLU \\
\hline
Reward discount factor  & 0.95 \\
\hline
GAE  & 0.95\\
\hline
PPO clipping  & 0.2  \\
\hline
Entropy temperature   & 0.2  \\
\hline
Mini-batch size & 64 \\
\hline
Learning rate & 0.0003 \\
\hline
Replay buffer size  & 5000 \\
\hline
Target network update coefficient   & 0.005 \\
\hline
FedAvg frequency &  4  \\
\hline
Epochs & 10 \\
\hline
SAC Critic networks & 2 \\
\hline
Gradient clipping & 0.5 \\
\hline
Policy update frequency & 40\\
\hline
Training episodes & 330 \\
\hline
QoE scene quality coefficient & 1 \\
\hline
QoE choppiness penalty coefficient & 0.4 \\
\hline
QoE latency penalty coefficient & 0.2 \\
\hline
QoE stability penalty coefficient & 0.6 \\
\hline
QoE disruption penalty coefficient & 0.5 \\
\hline
\end{tabular}
}
\end{table}

\begin{figure}[htbp]
\centering
\includegraphics[width=0.48\textwidth]{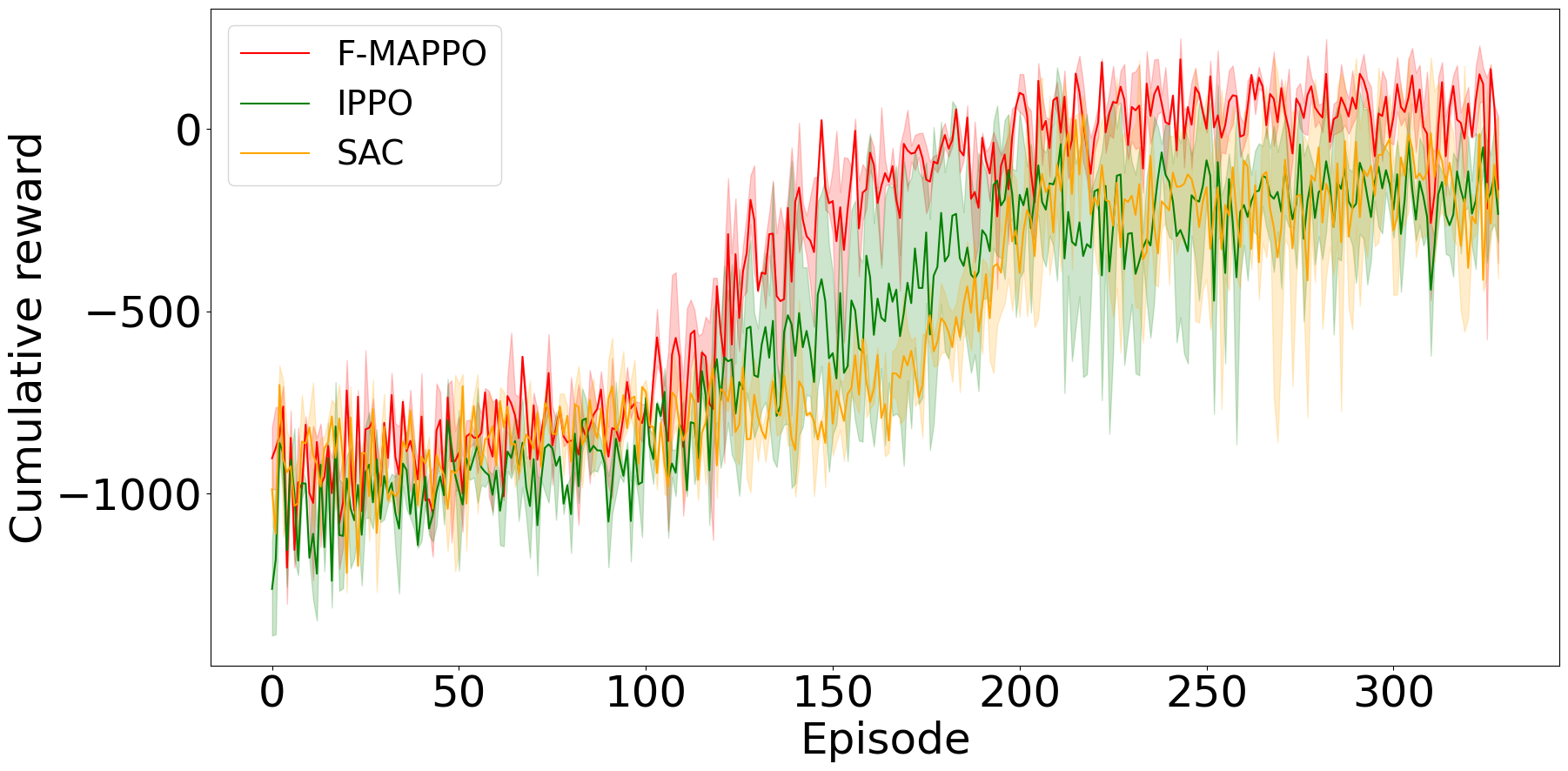}
\caption{Learning curves of the three DRL-based metaverse streaming methods: F-MAPPO, IPPO, and SAC over 330 episodes (40s for an episode) during the offline training.} 
\label{training}
\end{figure}

Unlike conventional DRL problems, the concept of an ``episode" in the metaverse lacks a well-defined boundary due to its continuous and dynamic nature. To address this, we empirically define an episode as a fixed 40-second interval, during which agents make decisions every second. To ensure adaptability to the ever-changing metaverse environment, we train each method across diverse network conditions (shown in Table. \ref{network_conditions}) over 12,000 time steps. The training processes, in terms of cumulative reward, are visually presented in Fig. \ref{training}. During the early stages, the cumulative rewards of all three methods remain relatively low, with minimal differences among them. This is primarily because the DRL models require extensive interactions with the environment to develop a reliable and efficient bit rate selection policy. After approximately 200 training episodes, the rewards stabilize, indicating policy convergence. Notably, F-MAPPO achieves the highest cumulative rewards, suggesting that it optimizes the QoE function more effectively, ultimately delivering the best user experience within dynamic social metaverse streaming system.

\subsection{Online Testing}

\begin{figure*}[htbp]
\centering
\includegraphics[width=0.95\textwidth]{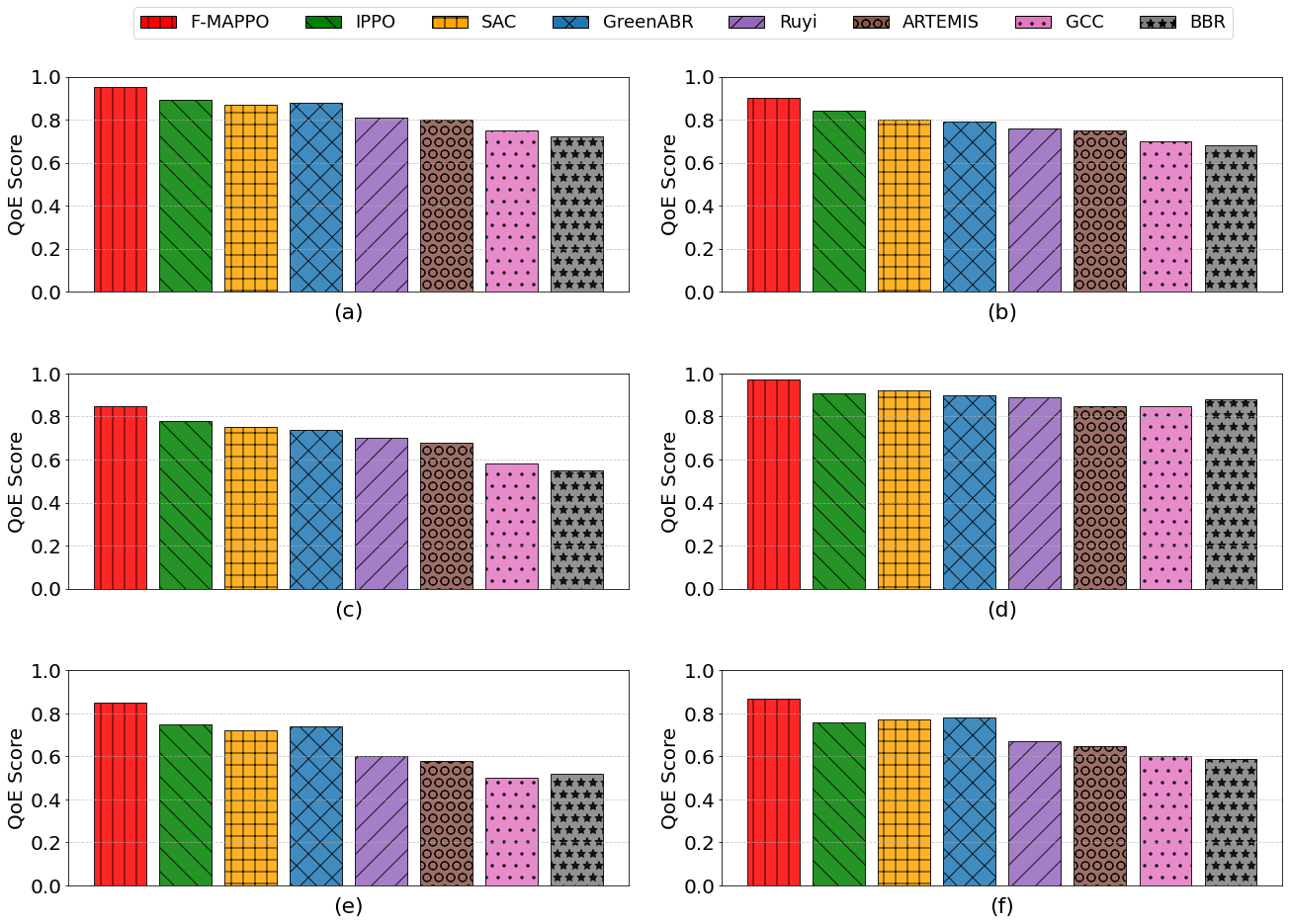}
\caption{Comparing F-MAPPO with other streaming methods under different network conditions: (a) high-performance cloud streaming, (b) home Wi-Fi 6 network, (c) 4G LTE mobile network, (d) 5G edge computing, (e) network congestion, and (f) network recovery.}
\label{compare}
\end{figure*}

To evaluate the performance of F-MAPPO in social metaverse streaming, we conducted extensive online testing under diverse network conditions. The experiments were designed to compare F-MAPPO against seven state-of-the-art adaptive streaming methods: IPPO, SAC, GreenABR, Ruyi, ARTEMIS, GCC, and BBR. As shown in Fig. \ref{compare}, F-MAPPO consistently achieved the highest QoE scores across all network conditions, outperforming the baseline methods in every scenario. In high-bandwidth, low-latency environments such as S1 and S4, F-MAPPO achieved QoE scores of 0.95 and 0.97, respectively, surpassing IPPO (0.89, 0.91), SAC (0.87, 0.92), and GreenABR (0.88, 0.90). While these methods performed well in ideal conditions, GCC and BBR struggled significantly, scoring only 0.75 and 0.72 in S1 and 0.85 and 0.88 in S4. 

In lower-bandwidth and higher-latency environments like S3 and S2, the performance gap between methods widened. F-MAPPO maintained its advantage, scoring 0.85 in S3 and 0.90 in S2, outperforming IPPO (0.78, 0.84) and SAC (0.75, 0.80). GreenABR, Ruyi, and ARTEMIS showed moderate performance, with QoE scores ranging from 0.68 to 0.79. However, GCC and BBR experienced sharp declines under these conditions, dropping to 0.58 and 0.55 in the LTE environment. The high latency (50–100 ms) and fluctuating bandwidth (20–80 Mbps) in S3 had a particularly strong impact on GCC and BBR. These conditions caused frequent buffering events and reduced frame stability, exposing their inefficiency in handling unstable wireless networks.

When network congestion (S5) occurred, all methods suffered a QoE drop, but F-MAPPO remained the most stable, maintaining a QoE of 0.85, while IPPO, SAC, and GreenABR showed greater fluctuations. Ruyi and ARTEMIS, which prioritize energy efficiency and adaptive resource allocation, struggled more in congestion scenarios. GCC and BBR exhibited the worst performance, with QoE values dropping below 0.55. This result showed their lack of adaptability to sudden bandwidth limitations. Conversely, during network recovery (S6), F-MAPPO demonstrated faster adaptation than other methods, reaching a QoE of 0.87, compared to 0.76–0.78 for IPPO, SAC, and GreenABR. GCC and BBR remained the slowest to recover, reinforcing their limitations in dynamic metaverse environments.

The results demonstrate that F-MAPPO consistently outperforms other baseline methods across all tested network conditions. In high-bandwidth, low-latency environments, it effectively utilizes available resources to deliver a more stable and high-quality streaming experience, minimizing bitrate fluctuations and frame drops. While other learning-based methods (IPPO, SAC, GreenABR, and Ruyi) adapt well in optimal conditions, they show instability when bandwidth availability fluctuates. Traditional methods (ARTEMIS, GCC, and BBR), in contrast, struggle significantly under these conditions, failing to fully utilize network capacity and leading to degraded user experiences. In scenarios with limited bandwidth and higher latency, such as 4G LTE and home Wi-Fi, F-MAPPO continued to outperform other methods, showing greater resilience against fluctuating network conditions. Other learning-based methods experienced noticeable performance drops, struggling to dynamically adjust to bandwidth constraints. When faced with sudden network fluctuations, such as congestion and recovery phases, F-MAPPO maintained its advantage by adapting faster and more efficiently than any other method. While other learning-based methods exhibited slower stabilization, traditional methods showed significant limitations in handling rapid bandwidth changes. These results confirm that F-MAPPO’s federated multi-agent framework provides a more robust solution for real-time streaming, making it the most effective choice for social metaverse applications that require high adaptability and consistent performance.

\subsection{Feasibility Analysis}
To measure the computational and communication overheads of FL, We trained both centralized MAPPO and F-MAPPO under the same network conditions. For computational overhead, GPU utilization averaged 68.2\% for centralized MAPPO and increased to 74.5\% for F-MAPPO due to the FL training process such as local model training, periodic aggregation, and secure updates. In terms of communication overhead, F-MAPPO only transmits local model gradients to the global agent server for aggregation. The size of each update is approximately 0.5 MB per round per device, leading to a total communication overhead of approximately 3 MB per aggregation round for 6 devices. Compared to centralized MAPPO, which transmits state-action data (around 1 KB per episode), the additional communication cost in F-MAPPO is negligible in practical streaming scenarios.

\section{Conclusion} \label{conclusion}
% In this paper, we propose ASMS, an adaptive social metaverse streaming system designed to enable multiple users to experience high-quality, remotely-rendered social metaverse scenes. To optimize the overall user experience, ASMS employs F-MAPPO, which utilizes multiple agents to dynamically select appropriate streaming bit rates for users under varying network conditions. Furthermore, all agents in F-MAPPO use a shared global model within a FL-based distributed framework, addressing the privacy issues inherent in centralized learning. Experimental results have demonstrated the effectiveness and advantages of F-MAPPO across all tested scenarios, highlighting its superior performance in optimizing user experience while preserving data privacy. 

In this paper, we propose ASMS, an adaptive social metaverse streaming system that enables multiple users to experience high-quality, remotely rendered metaverse scenes. To optimize the overall user experience, ASMS employs F-MAPPO, a framework that allows multiple agents to dynamically select streaming bit rates under varying network conditions. Unlike conventional ABR methods, which optimize streaming for a single user, F-MAPPO applies MADRL to ensure efficient and fair resource allocation in multi-user metaverse environments. All agents in F-MAPPO use a shared global model within a FL-based distributed framework. This avoids the limitations of existing methods that rely on centralized data collection for training, which often raises privacy concerns and requires direct access to raw user data. By keeping data on local devices and sharing only model updates, ASMS enhances privacy while maintaining learning efficiency. Future work will focus on accelerating model convergence to improve training efficiency and real-time adaptability to support large-scale users in dynamic metaverse environments.

% or


% ============================================
%\appendices
%Appendix one text goes here %\cite{Roberg2010}.

%\section{}

% \section*{Acknowledgment}
% The author Zijian Long wants to thank China Scholarship Council (CSC) for the financial support.

%The authors would like to thank D. Root for the loan of the SWAP. The SWAP that can ONLY be usefull in Boulder...

\ifCLASSOPTIONcaptionsoff
  \newpage
\fi

% The ``triggered" command can be changed if desired:


%\begin{thebibliography}{1}


\bibliographystyle{IEEEtran}
\bibliography{Bibliography}
%\end{thebibliography}
%\begin{biography}[{\includegraphics[width=1in,height=1.25in,clip,keepaspectratio]{mshell}}]{Michael Shell}

% ==== SWITCH OFF the BIO for submission
% ==== SWITCH OFF the BIO for submission

\vskip -25pt plus -1fil
\begin{IEEEbiography}[{\includegraphics[width=1in,height=1.25in,clip,keepaspectratio]{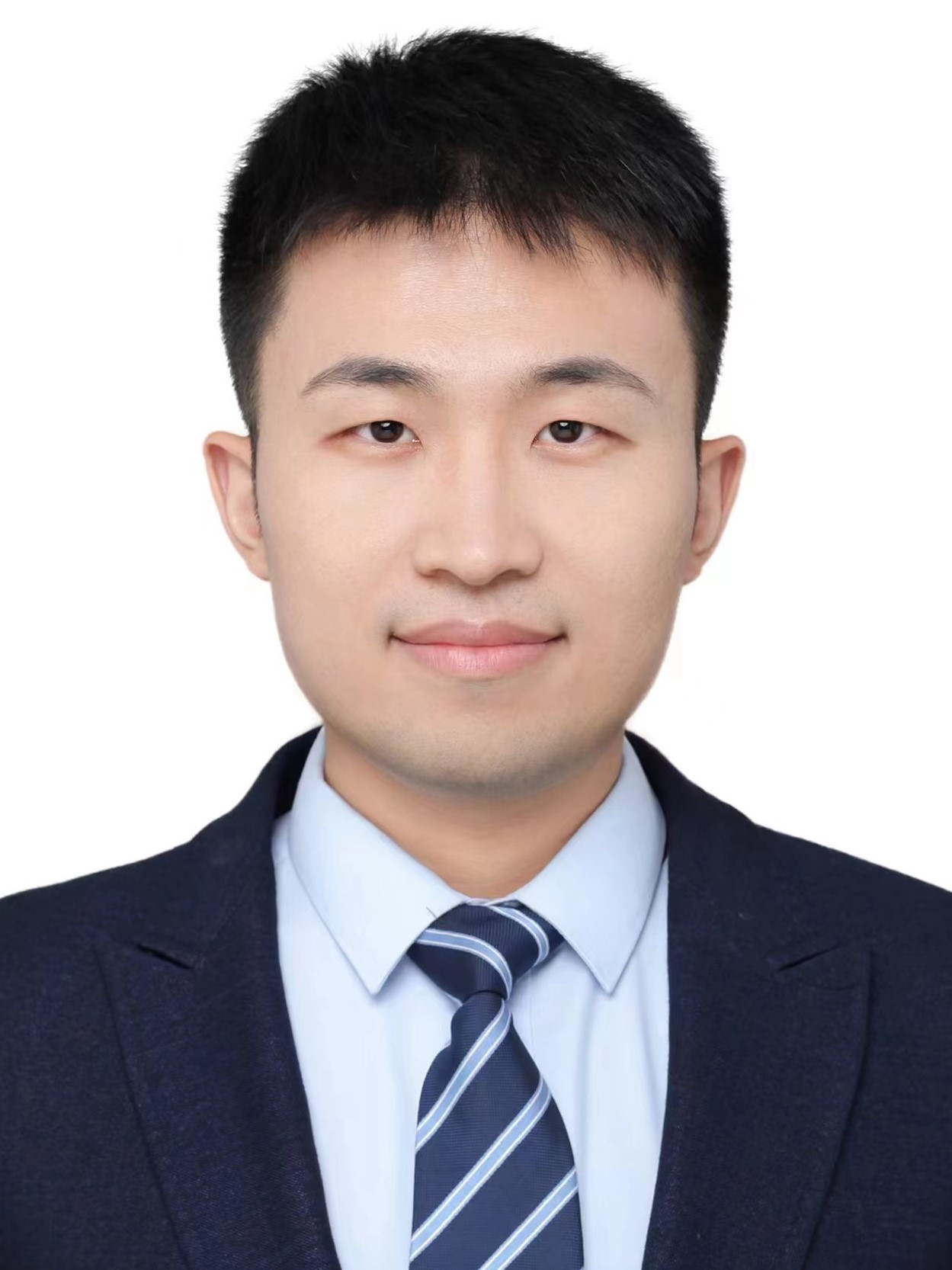}}]{Zijian Long}
(zlong038@uottawa.ca) received the B.Sc. degree in Software Engineering from Beijing Institute of Technology, China, in 2016 and the M.Sc. degree in Electrical and Computer Engineering from the University of Ottawa, Canada, in 2020. He is currently a Ph.D. candidate in the School of Electrical Engineering and Computer Science, University of Ottawa. His research interests include metaverse, XR network, and reinforcement learning.
\end{IEEEbiography}

\vspace{-8 mm}

\begin{IEEEbiography}[{\includegraphics[width=1in,height=1.25in,clip,keepaspectratio]{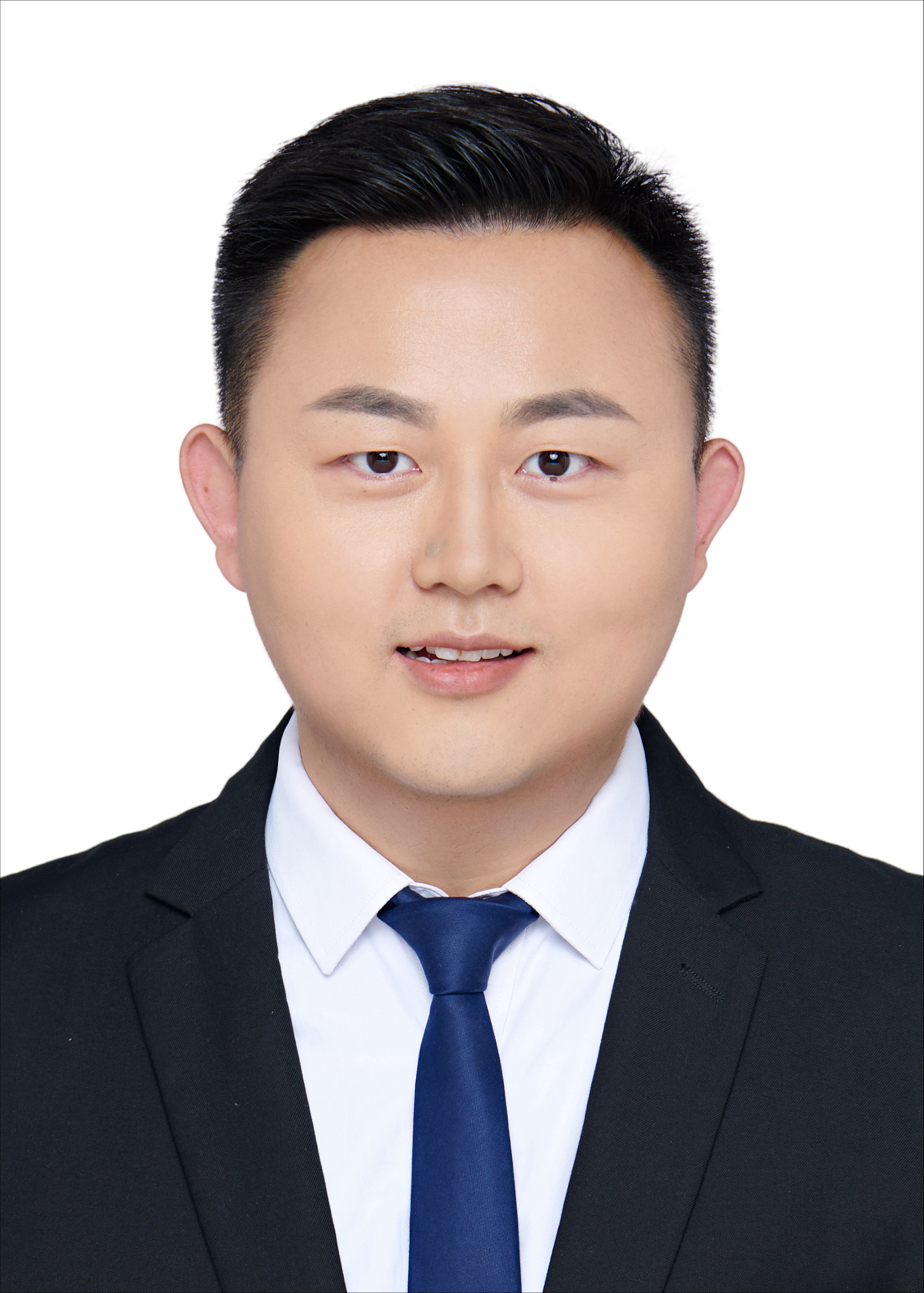}}]{Haopeng Wang}(hwang266@uottawa.ca) received the M.Eng. degree in electronic and communication engineering and B.Eng. degree in information and electronics from Beijing Institute of Technology, Beijing, China, in 2017 and 2015, respectively. He is currently pursuing the Ph.D. degree in electrical and computer engineering at the University of Ottawa. His research interests are AI, computer network, extended reality, and multimedia.
\end{IEEEbiography}

\vspace{-8 mm}

\begin{IEEEbiography}[{\includegraphics[width=1in,height=1.25in,clip,keepaspectratio]{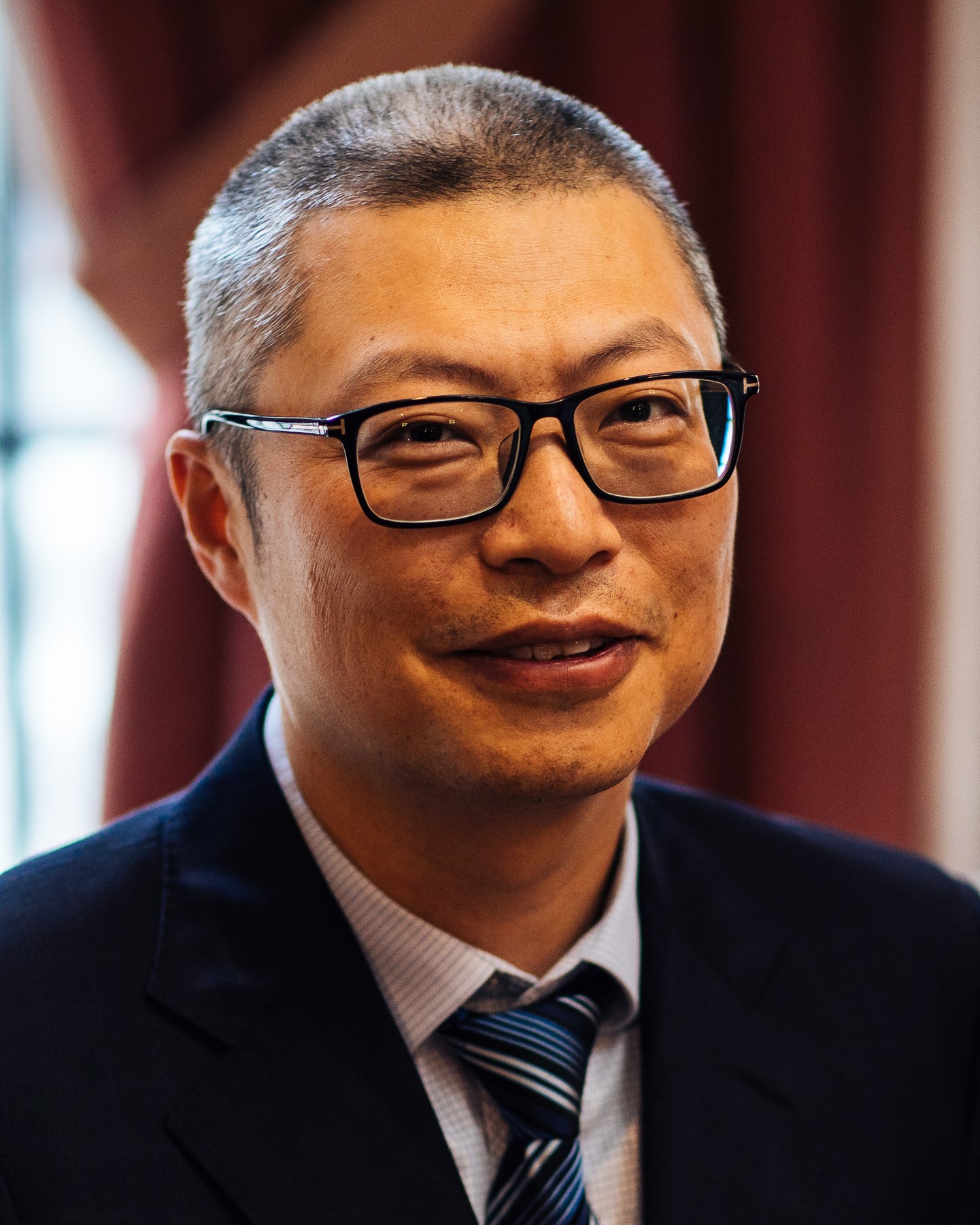}}]{Haiwei Dong}
(haiwei.dong@ieee.org) is a Director and  Principal Researcher with Huawei Canada, and an Adjucnt Professor with the University of Ottawa. He was a Principal Engineer with Huawei Canada, Toronto, ON, Canada, a Research Scientist with the University of Ottawa, Ottawa, ON, Canada, a Postdoctoral Fellow with New York University, New York, NY, USA, a Research Associate with the University of Toronto, Toronto, ON, Canada, and a Research Fellow (PD) with the Japan Society for the Promotion of Science, Tokyo, Japan. He received the Ph.D. degree from Kobe University, Kobe, Japan in 2010 and the M.Eng. degree from Shanghai Jiao Tong University, Shanghai, China, in 2008. He also serves as a Department Editor of IEEE Multimedia Magazine; an Associate Editor of ACM Transactions on Multimedia Computing, Communications and Applications; and an Associate Editor of IEEE Consumer Electronics Magazine. He is a Senior Member of IEEE, a Senior Member of ACM, and a registered Professional Engineer in Ontario. His research interests include artificial intelligence, multimedia, metaverse, digital twin, and robotics.
\end{IEEEbiography}

\vspace{-8 mm}

\begin{IEEEbiography}[{\includegraphics[width=1in,height=1.25in,clip,keepaspectratio]{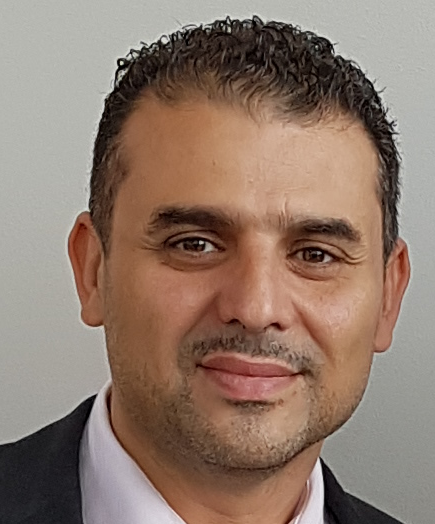}}]{Abdulmotaleb El Saddik}
(elsaddik@uottawa.ca) is currently a Distinguished Professor with the School of Electrical Engineering and Computer Science, University of Ottawa. He has supervised more than 150 researchers. He has coauthored ten books and more than 650 publications and chaired more than 50 conferences and workshops. His research interests include the establishment of digital twins to facilitate the well-being of citizens using AI, the IoT, AR/VR, and 5G to allow people to interact in real time with one another as well as with their smart digital representations. He received research grants and contracts totaling more than \$30 M. He is a Fellow of Royal Society of Canada, a Fellow of IEEE, an ACM Distinguished Scientist and a Fellow of the Engineering Institute of Canada and the Canadian Academy of Engineers. He received several international awards, such as the IEEE I\&M Technical Achievement Award, the IEEE Canada C.C. Gotlieb (Computer) Medal, and the A.G.L. McNaughton Gold Medal for important contributions to the field of computer engineering and science.
\end{IEEEbiography}

%% insert where needed to balance the two columns on the last page with
%%\newpage

% ==== SWITCH OFF the BIO for submission
% ==== SWITCH OFF the BIO for submission


% \vfill


\end{document}